# Beyond Imitation: Auditing the Recoverability of Reasoning in Distilled Models

Ruitong Li[1,*] Binjie Guo[2,*] Aisheng Mo[2] Guowei Su[2]
Han Wang[3] Jie Li[4] Ru Zhang[2]

[1]University of Hong Kong [2]Zhejiang University
[3]Dalian University of Technology [4]Independent Researcher

[*]These authors contributed equally.

**Abstract**

A correct teacher solution becomes useful supervision when the receiving student can continue its reasoning. We measure this compatibility with *prefix recovery*: after revealing 25%, 50%, or 75% of a verified solution, we test whether the student completes it correctly. We connect recovery to the cosine conflict between cross-entropy and reverse-KL gradients over the full vocabulary. Across adjacent Qwen3 teacher–student pairs from 0.6B to 8B parameters, reverse-KL distillation delivers its most consistent mathematical and code improvements for the two students below 2B parameters. On a fixed cohort of 1,000 trajectories, average prefix recovery rises from 71.0% to 91.9% as student size increases from 0.6B to 4B, and the robust–fragile recovery gap contracts from 46.0 to 14.4 percentage points. With the teacher fixed at 8B, conflict separation falls from 0.993 to 0.233. An independent objective intervention finds the largest reverse-KL rescue on fragile trajectories. The three measurements locate the same capacity-dependent transfer regime: distribution matching has the greatest headroom when correct traces remain unevenly recoverable. Prefix recovery provides a practical diagnostic for selecting costly distribution-level distillation.

## 1 Introduction

Knowledge distillation transfers predictive structure from a teacher to a smaller student (Buciluǎ et al., 2006; Hinton et al., 2015). For autoregressive language models, sequence supervision (Kim and Rush, 2016), reverse-KL distribution matching (Gu et al., 2024), and on-policy training (Agarwal et al., 2024) improve compact generators. Reasoning distillation additionally transfers natural-language rationales (Hsieh et al., 2023; Li et al., 2023). These successes motivate an assumption: once a teacher produces a correct reasoning trace, stronger teachers should provide better supervision.

Correctness leaves student compatibility unresolved. Distillation can have an optimal teacher scale (Zhang et al., 2025); small models can struggle with traces generated by stronger reasoners (Li et al., 2025); and teacher selection remains an open problem (Panigrahi et al., 2026). Existing studies characterize this mismatch through aggregate teacher scale, trace complexity, or downstream validation. The missing quantity is whether a specific student can use the intermediate states of a specific correct solution.

Prefix recovery measures this student–trajectory compatibility. For each verified trajectory, we reveal a controlled fraction of its reasoning and sample student continuations. Their success rate defines trajectory recoverability for that student. We also compare the token-level directions induced by one-hot cross-entropy and full-distribution reverse KL, defining their cosine disagreement as *objective conflict*. High conflict marks states where the two supervision signals move a capacity-limited student in different directions. Figure 1 summarizes the design.

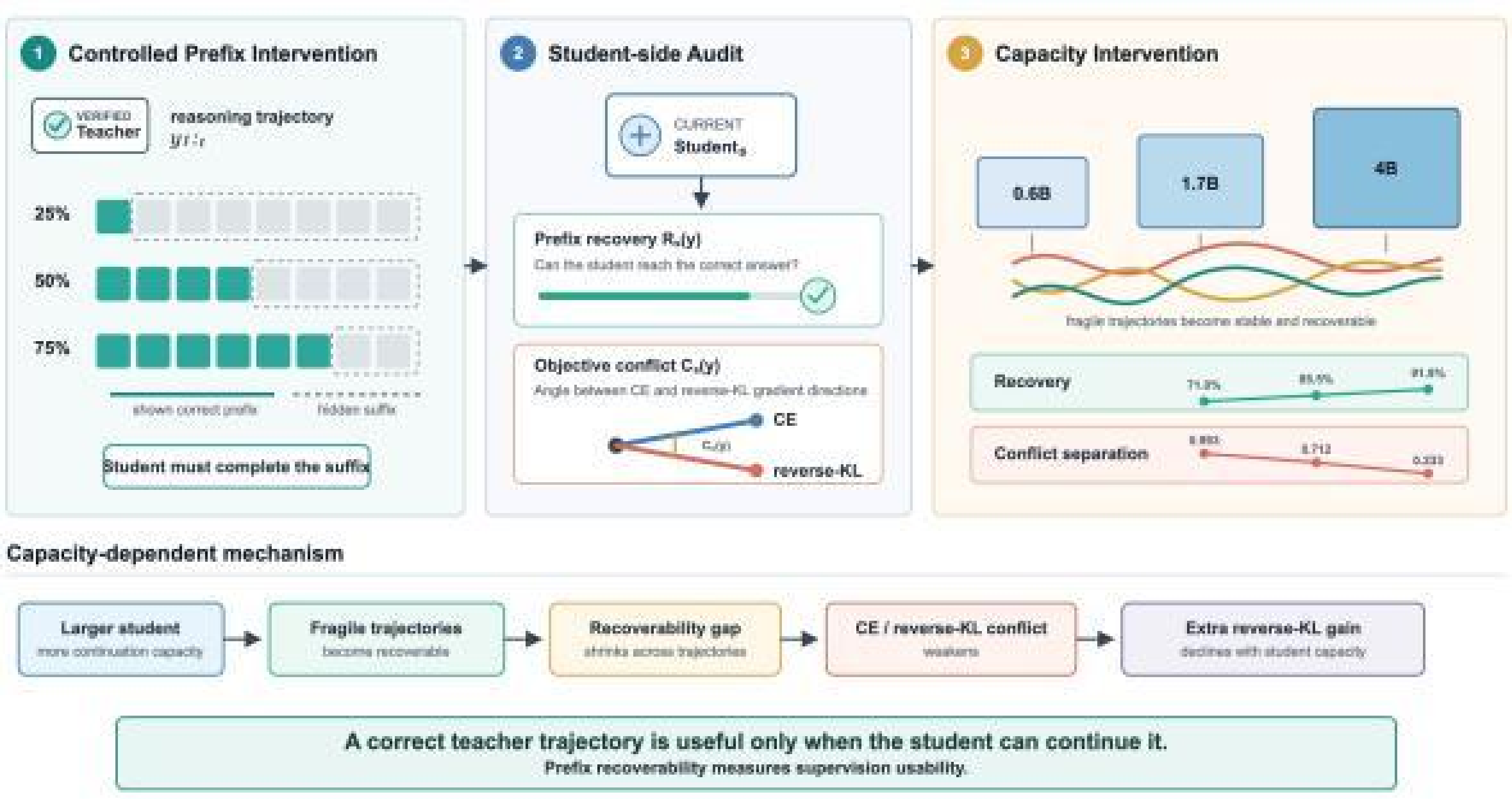

Figure 1: Student-aware trajectory distillation through prefix recoverability. Controlled prefix interventions measure whether a student can continue a verified teacher trajectory; exact CE–reverse-KL conflict measures disagreement between supervision directions. Repeating the audit across student capacities links rising recoverability to weaker conflict and diminishing additional reverse-KL gain.

The experiments produce three results. First, adjacent-scale rKL raises OlympiadBench and HumanEval accuracy by 4.30 and 3.66 points for the 0.6B student. The 1.7B student gains 2.58 and 2.44 points. Second, mean prefix recovery rises from 71.0% at 0.6B to 91.9% at 4B, and the robust–fragile difference shrinks by 31.6 percentage points. Third, with the teacher fixed at 8B, standardized objective-conflict separation decreases from 0.993 to 0.712 and 0.233. Downstream transfer, behavioral recovery, and optimization geometry identify the same capacity trend.

Our contribution is an intervention-based account of distribution-transfer headroom. Reverse KL helps most when correct trajectories differ sharply in recoverability and that difference aligns with objective conflict. As capacity grows, fragile trajectories become recoverable, conflict loses predictive strength, and the additional gain from teacher matching diminishes.

## 2 Related Work

Classical knowledge distillation transfers softened teacher outputs (Hinton et al., 2015). Sequence targets (Kim and Rush, 2016), born-again students (Furlanello et al., 2018), teacher assistants (Mirzadeh et al., 2020), and self-distillation (Mobahi et al., 2020) extend this interface. Language-model compression also aligns intermediate representations (Sun et al., 2019; Jiao et al., 2020), self-attention relations (Wang et al., 2020), and pretraining objectives (Sanh et al., 2019). Each interface carries a compatibility requirement between teacher and student. Hidden-state matching relies on correspondence between layers; relation matching transfers internal geometry; response-level matching exposes the student to teacher token distributions and sequence states. These requirements become consequential for reasoning, where one verified answer can contain several transformations with different demands on the receiving model. Aggregate benchmark accuracy merges trajectory compatibility with optimization, data coverage, and raw task ability. Teacher assistants reduce model-level capacity gaps and reveal their value after training. Prefix recovery evaluates compatibility before training by testing whether the receiving student can continue a fixed teacher trajectory. Its unit of analysis is the student–trajectory pair, allowing the same verified trace to be usable for one student and fragile for another. This focus complements broader compression work on pruning, quantization, and distillation (Zhu et al., 2024).

Generative distillation increasingly exposes the teacher distribution or the student's state distribution. MiniLLM uses reverse KL to suppress probability mass outside teacher support (Gu et al., 2024); GKD trains on student-generated contexts with flexible divergences (Agarwal et al., 2024); and recent work studies KL direction and transport-based objectives (Wu et al., 2025; Cui et al., 2024). Objective geometry determines which errors receive weight. Teacher-forced traces supply a high-quality target path. Student-generated contexts capture familiar states. Prefix intervention supplies a third view: it enters a verified teacher path at a controlled point and observes the student's conditional continuation. Reasoning methods contribute another interface. Rationale distillation transfers explanations (Hsieh et al., 2023; Li et al., 2023), chain-of-thought prompting and STaR construct reasoning paths (Wei et al., 2022; Zelikman et al., 2022), and process supervision evaluates intermediate steps (Lightman et al., 2024). Verified traces and process feedback now support strong mathematical reasoners (Shao et al., 2024; Guo et al., 2025). Step validity establishes a sound teacher path; prefix recovery measures its usability for the recipient. Objective conflict connects that usability to the update directions produced by one-hot and distribution-level supervision.

Capacity mismatch provides the closest context. Teacher assistants insert an intermediate model (Mirzadeh et al., 2020); scaling studies find capacity-dependent teacher optima (Zhang et al., 2025); compact reasoners struggle with some traces from stronger models (Li et al., 2025); and gradient-based methods select teachers whose updates suit the student (Panigrahi et al., 2026). Weak-to-strong generalization likewise separates supervisor accuracy from supervision quality (Burns et al., 2024). Existing selection signals operate at several resolutions. Teacher-level scores summarize an entire generator. Example difficulty combines the problem with the chosen reasoning route. Static confidence measures surprise at one state. Prefix recovery resolves the analysis to individual student–trajectory pairs and actively varies the amount of valid context. Repeating the same intervention across students tracks a trace from fragility at 0.6B through partial repair at 1.7B to reliable continuation at 4B. The mechanism predicts faster recovery for fragile trajectories, contraction of the recovery gap, and weaker objective disagreement on the same trajectory identities. A fixed-teacher factorial isolates receiving capacity, and cross-domain evaluation connects the trajectory mechanism to downstream transfer.

# 3 Prefix Recovery and Objective Conflict

## 3.1 Measuring Trajectory Recoverability

Let $y = (y_1, ..., y_T)$ be a verified correct trajectory and let $f \in \{.25, .50, .75\}$. We reveal $y_{1:\lfloor fT \rfloor}$ to student $s$ and sample $K = 4$ continuations. With verifier $V$, prefix recovery is

$$R_s(y) = \frac{1}{3K} \sum_{f} \sum_{k=1}^{K} V\left(y_{1:\lfloor fT \rfloor} \oplus \hat{y}_{f,s}^{(k)}\right). \tag{1}$$

Unlike confidence or trace length, $R_s(y)$ changes the amount of correct reasoning available to the student. Low recovery means that even a verified trajectory offers little usable support; high recovery means that partial reasoning reliably places the student on a successful continuation path. Because $R_s$ is indexed by the student, recoverability can change with capacity.

Recovery assigns a behavioral meaning to every student–trajectory pair. The verifier evaluates the completed answer, and the controlled prefix determines how much valid reasoning enters the context. Averaging K continuations estimates success under the fixed decoding policy. Repeating the procedure at three fractions distinguishes an early transition failure from a path that remains difficult near its end. The resulting statistic can be computed before distillation and reused across candidate objectives.

The three prefix fractions form a controlled dose response. Failure after a short prefix can reflect an inaccessible early transition or insufficient task information. Success after a long prefix localizes the failure: the student can perform the terminal computation once enough of the reasoning path is supplied. Failure at every fraction places the demonstrated path outside the student's continuation support. Their average gives one recovery statistic per trajectory, and the fraction-specific values retain the dose-response shape. Reusing identical trajectories and prefix fractions across capacities attributes changes in recovery to the receiving model under the frozen protocol.

## 3.2 Capacity–Matched Reverse KL

For student $p_\theta$ and frozen teacher q, the transfer arm minimizes full-vocabulary reverse KL on response tokens:

$$\mathcal{L}_{\mathrm{rKL}}(x, y) = \sum_{t \in \mathcal{R}} D_{\mathrm{KL}}(p_\theta(\cdot \mid x, y_{<t}) \,\|\, q(\cdot \mid x, y_{<t})) \,. \tag{2}$$

Our capacity intervention uses four adjacent pairs: 0.6B←1.7B, 1.7B←4B, 4B←8B, and 8B←14B. This progression changes the receiving capacity under one fixed reverse-KL objective.

Reverse KL communicates the shape of teacher support at every response token. Plausible alternatives receive structured probability even when the recorded trajectory contains one realization. This information is most consequential at states where the student has a narrow successful continuation set. The adjacent-scale progression measures how that opportunity changes as the recipient gains capacity under the same transfer interface.

## 3.3 Exact Objective Conflict

At a response token, let $g_{\mathrm{CE}} = \nabla_z[-\log p_\theta(y_t)]$ and $g_{\mathrm{rKL}} = \nabla_z D_{\mathrm{KL}}(p_\theta \| q)$ for student logits z. We define

$$C_t = \frac{1 - \cos(g_{\mathrm{CE}}, g_{\mathrm{rKL}})}{2}, \qquad C(y) = \frac{1}{|\mathcal{R}|} \sum_{t \in \mathcal{R}} C_t. \tag{3}$$

For $p_i = p_\theta(i \mid x, y_{<t})$, $q_i = q(i \mid x, y_{<t})$, and $D = D_{\mathrm{KL}}(p \| q)$, the reverse-KL logit gradient is

$$\frac{\partial D}{\partial z_i} = p_i \left(\log p_i - \log q_i - D\right) . \tag{4}$$

CE privileges the verified next token, whereas reverse KL redistributes probability across the full teacher support. High conflict identifies states where those local directions disagree. This yields three predictions: student capacity should increase recovery, reduce the robust–fragile recovery gap, and weaken conflict–recovery coupling; downstream rKL gains should diminish in the same regime.

The normalization in Equation 3 maps aligned gradients to zero and opposing gradients to one. Averaging over valid response tokens produces a trajectory-level quantity that shares identities with the recovery audit. We compare its robust–fragile separation and its rank association with recovery. Separation tests whether fragile trajectories induce systematically different supervision geometry; rank association tests whether that geometry orders trajectory usability within a student scale.

Objective conflict is evaluated at the same states as prefix recovery. Low recovery with low conflict indicates difficulty shared by both objectives. Low recovery with high conflict identifies a trajectory where one-hot imitation concentrates on the realized token and reverse KL preserves teacher-supported alternatives. The distributional update can provide a smoother local target for a capacity-limited student. This mechanism predicts larger reverse-KL gains on fragile trajectories and a contracting advantage as capacity makes those states reachable.

## 4 Why Capacity Changes Distribution Transfer

Cross-entropy and reverse KL expose different parts of a correct trajectory. CE supplies one privileged token at every position. Reverse KL compares the complete student and teacher distributions, preserving locally interchangeable computations and discouraging unsupported branches. Their gradients diverge when matching teacher support competes with increasing the probability of the recorded token.

The value of this distributional information depends on the student's reachable continuation set. A small student may reach the answer through a narrow set of continuations, making its few successful paths sensitive to probability shifts. A larger student can represent several successful continuations and absorb broader teacher support. Prefix recovery measures this distinction directly from continuation behavior.

At a fragile state, the recorded token points toward one successful continuation and the teacher distribution assigns mass to several locally valid alternatives. CE amplifies the recorded branch. Reverse KL shapes the neighborhood around that branch. Their disagreement becomes useful when the student's own distribution is concentrated on brittle or unsupported continuations. Once several successful branches enter the student's reachable set, the same distributional information produces a smaller behavioral change.

This view predicts a specific form of diminishing return. Capacity first repairs low-recovery trajectories, raising average recovery and contracting variation across trajectories. Objective conflict then loses its association with failure, and full-distribution transfer has less room to improve the raw student. The prediction couples trajectory-level recovery, token-level gradient geometry, and item-level downstream accuracy.

This ordered mechanism defines the capacity boundary operationally. Below the boundary, recovery varies widely across verified traces and objective conflict tracks that variation. Distribution matching has identifiable states to repair. Near the boundary, fragile traces become recoverable and the conflict–recovery association approaches zero. Additional reverse-KL gains contract at the same scales. The boundary is measured through these linked changes, giving it a behavioral and optimization interpretation.

Three controls identify the source of attenuation. A fixed-teacher comparison isolates receiving capacity from teacher identity. Training-trajectory recovery and conflict separate the mechanism from benchmark saturation. Multiple prefix fractions measure a dose response beyond static solution length and confidence.

## 5 Predictions and Identification

We organize the study around three linked predictions. First, increasing student capacity should raise recovery most strongly on the fragile cohort, causing the robust–fragile difference to contract. Second, exact CE–rKL conflict should be larger on fragile trajectories and its separation and rank association with recovery should weaken as students grow. Third, downstream rKL gains should be largest in the capacity range where the first two signals remain strong. Each prediction is evaluated at its natural experimental unit: trajectories for recovery, response tokens for conflict, and benchmark items for transfer.

The robust and fragile cohorts are fixed before the cross-scale comparison. They contain the same trajectory identities at 0.6B, 1.7B, and 4B, so cohort movement records how larger recipients repair states identified by the smallest student. Length matching keeps the comparison focused on continuation behavior. The primary recovery contrast is the difference between cohort means at each scale, followed by its contraction across scales.

Within-trajectory pairing gives every scale the same verified reasoning. The fixed-8B factorial changes the student and preserves teacher identity. Mathematics and code provide domain replication through disjoint sources, verifiers, and output structures. We report each domain separately to preserve these distinct outcomes.

The fixed-teacher factorial adds a second axis to the design. Student capacity changes across 0.6B, 1.7B, and 4B under the same 8B teacher. Separate fixed-student comparisons change the teacher from 4B to 8B. Stable trajectory rankings across teachers, combined with weakening conflict across students, assign the dominant ordering of difficult trajectories to the recipient. Domain replication then tests whether this student-side mechanism accompanies transfer in both symbolic mathematics and executable code.

The estimand is the joint change in recoverability, objective conflict, and transfer under one frozen training protocol. The fixed-teacher comparison and paired trajectories identify how receiving capacity participates in the observed mechanism.

# 6 Experimental Protocol

We use dense Qwen3 models (Yang et al., 2025). Mathematical transfer trains on GSM8K (Cobbe et al., 2021) trajectories and evaluates on 581 English competition-math problems from OlympiadBench (He et al., 2024). Code transfer trains on MBPP (Austin et al., 2021) and evaluates on 164 HumanEval problems (Chen et al., 2021). Every transfer arm follows the same 800-step budget with effective batch size 16. Decoding and verification remain fixed within paired comparisons. Appendix A reports the complete optimization and update configuration.

The recovery study reuses the same 1,000 verified trajectories at every scale, with 12 continuations per trajectory and student. Exact conflict is computed for five student–teacher pairs, including a fixed-8B-teacher comparison that isolates student capacity from teacher identity. Accuracy comparisons retain item correspondence; recovery comparisons remain paired by trajectory.

All reported differences preserve the experimental unit. Benchmark comparisons pair correctness on the same item. Recovery comparisons pair the same trajectory across capacities and prefix fractions. Conflict summaries average valid response tokens within a trajectory before computing cohort separation and Spearman association. This structure prevents multiple continuations or tokens from inflating the benchmark sample size. Complete split sizes, optimization settings, verifier rules, and artifact checks appear in the supplementary material.

# 7 Results

## 7.1 Distribution Matching Helps Most Below 2B

Table 1 shows a capacity-bounded transfer regime. The 0.6B student gains 4.30 points on OlympiadBench and 3.66 points on HumanEval. The 1.7B student gains 2.58 and 2.44 points. At 4B and 8B, the mathematical changes are 0.00 and −1.03 points; code gains contract to 1.83 and 0.61 points. Reverse KL improves both domains for the two evaluated students below 2B, with diminishing additional benefit at larger scales.

The scale ordering is consistent across domains at the point most relevant to the mechanism. The largest gain appears at 0.6B, the gain contracts at 1.7B, and the two larger students show little additional headroom. Mathematics reaches its plateau earlier; code retains a small positive increment. This domain difference motivates separate reporting and supplies a useful stress test: the shared conclusion rests on capacity ordering across tasks with different absolute accuracy and ceiling behavior.

The largest mathematical gain occurs at 0.6B. Separate domain reporting shows the replicated positive direction across mathematics and code at both sub-2B scales. The attenuation begins before either benchmark is solved: the 4B raw student reaches 48.02% on OlympiadBench and 73.78% on HumanEval. Absolute capability leaves substantial room for improvement as the same distribution-transfer interface contributes less. This separation between benchmark error and transfer headroom motivates the trajectory-level audit.

## 7.2 Trajectory Recoverability Rises with Capacity

The behavioral measurements follow the same capacity trend. Mean prefix recovery increases from 70.97% at 0.6B to 85.45% at 1.7B and 91.93% at 4B (Figure 2). The robust–fragile gap contracts from 45.97 to 25.83 and 14.35 percentage points, a paired reduction of 31.62 points from 0.6B to 4B. Larger students repair the trajectories that define the fragile cohort, reducing the headroom for distribution matching.

The paired recovery increase is 14.48 percentage points from 0.6B to 1.7B and 6.48 points from 1.7B to 4B. The smaller increment at 4B reflects trajectories approaching reliable recovery and matches the contraction in distillation gain.

The cohort decomposition locates the increase. Robust trajectories begin near reliable recovery and rise from 93.95% to 99.10%. Fragile trajectories move from 47.98% to 84.75% over the same capacity range. Most additional continuation ability is concentrated on trajectories unavailable to the smallest student. This asymmetric repair supplies the behavioral signature of the proposed mechanism.

The two cohorts play different roles in the transfer regime. Robust trajectories supply little differential signal because every recipient can already complete them from partial reasoning. Fragile trajectories expose the states where receiving capacity matters. Their 36.77-point recovery increase accounts for nearly all contraction of the cohort gap. Capacity expands usable support by repairing this specific part of the trajectory distribution, matching the location where reverse KL has its strongest objective-level advantage.

Controlled prefixes separate entry into a solution path from completion after the decisive transformation is available. Success after 75% of a trace and failure after 25% localizes the difficulty to an earlier transition. Failure at every fraction places the demonstrated continuation outside the student's support. Averaging the three intervention points gives a common estimand across scales, and fraction-specific results retain the response shape.

At 0.6B, the large robust–fragile difference identifies many correct traces with unrealized student support. Distribution matching can redistribute probability over alternative next steps at these states. By 4B, both cohorts approach reliable completion and the difference contracts to 14.35 percentage points. Fewer verified states distinguish usable from unusable supervision, producing the downstream plateau in additional transfer gain.

The mechanism requires recovery heterogeneity and CE–rKL conflict to weaken together as transfer gains contract. The next analysis tests this joint prediction with exact full-vocabulary gradients and a fixed teacher.

## 7.3 Objective Conflict Identifies the Capacity Boundary

Table 2 connects recovery to optimization. For the 0.6B student with a 4B teacher, fragile trajectories induce 0.0370 more objective conflict than robust trajectories, and conflict correlates with recovery at $\rho = -0.502$. With the teacher fixed at 8B, standardized conflict separation decreases from 0.993 to 0.712 and 0.233 as student capacity increases. The absolute conflict–recovery correlation falls from 0.524 to 0.085 (Figure 3). Fixed-student conflict rankings remain stable when teacher size changes from 4B to 8B ($\rho$ = 0.931 at 0.6B and 0.652 at 1.7B), assigning the attenuation primarily to the receiving student.

| | OlympiadBench | | | HumanEval | | |
|---|---|---|---|---|---|---|
| Student | Base | rKL | Δ | Base | rKL | Δ |
| 0.6B | 21.17 | **25.47** | **+4.30** | 33.54 | **37.20** | **+3.66** |
| 1.7B | 36.83 | **39.41** | **+2.58** | 57.93 | **60.37** | **+2.44** |
| 4B | 48.02 | 48.02 | +0.00 | 73.78 | **75.61** | +1.83 |
| 8B | 49.91 | 48.88 | -1.03 | 82.93 | **83.54** | +0.61 |

Table 1: Complete OOD accuracy (%) across student scales. Bold marks improvements over the undistilled student.

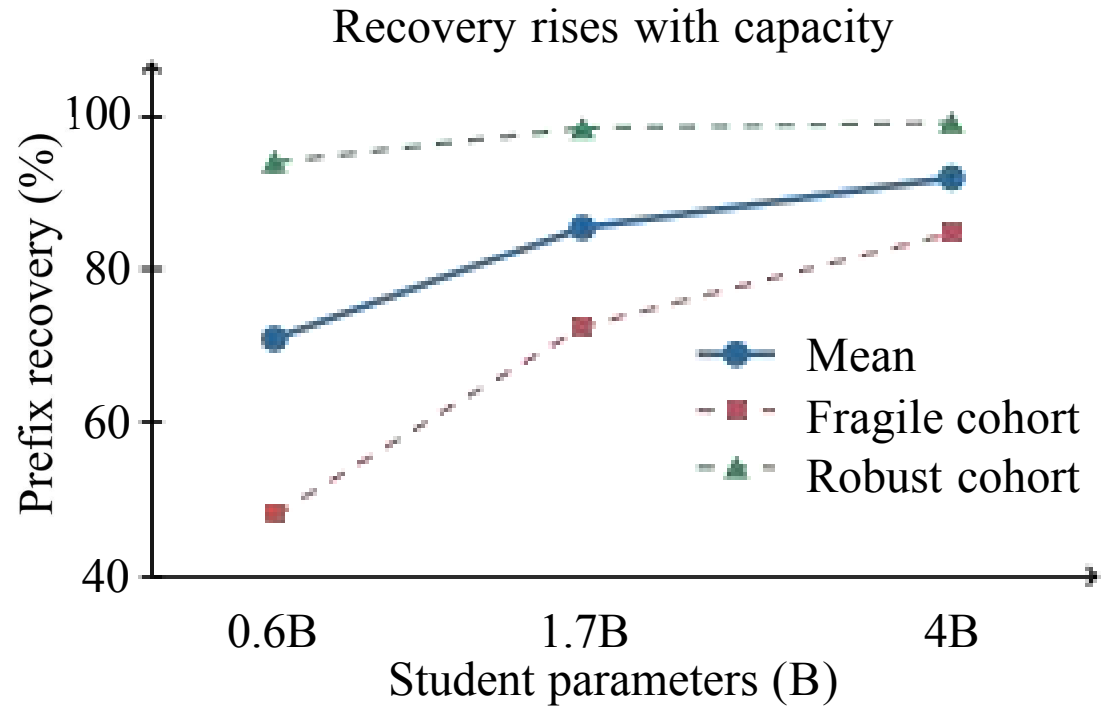

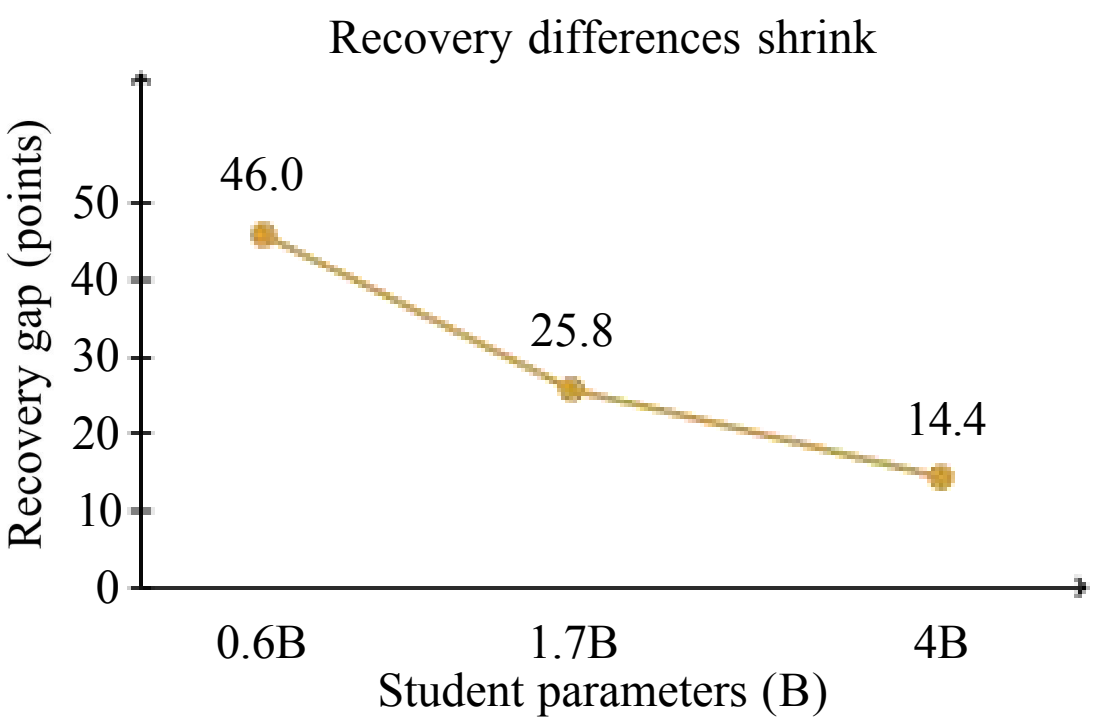

Figure 2: Prefix recovery on the same 1,000 verified trajectories. Average recovery rises with student capacity (left); the difference between robust and fragile cohorts shrinks (right).

| Student/teacher | Conflict gap | Std. effect | ρ(C, R) |
|---|---|---|---|
| 0.6B/4B | 0.0370 | 0.968 | -0.5016 |
| 0.6B/8B | 0.0446 | 0.993 | -0.5240 |
| 1.7B/4B | 0.0210 | 0.615 | -0.3092 |
| 1.7B/8B | 0.0234 | 0.712 | -0.3547 |
| 4B/8B | 0.0064 | 0.233 | -0.0848 |

Table 2: Exact full-vocabulary objective conflict. Positive gaps indicate greater CE–rKL disagreement on fragile trajectories.

An independent 2 $\times$ 2 objective intervention provides convergent evidence. On StepHard, rKL rescues 15.76 points over CE for fragile trajectories, compared with 8.79 points for stable trajectories. The 6.97-point interaction and a 4.25-point mean OOD rescue across ASDiv, SVAMP, and MATH-500 (Hendrycks et al., 2021) show that distribution matching helps most where one-hot imitation is behaviorally fragile.

This factorial design changes the training objective within each frozen cohort. The cohort contrast identifies trajectories with different measured recoverability; the objective contrast identifies the effect of adding full teacher support; their interaction asks whether distribution matching preferentially repairs fragile supervision. The larger reverse-KL gain in the fragile cell follows the direction predicted before the intervention. Transfer to three separate mathematical datasets shows that the repair extends beyond the trajectory source used to define the cohorts.

The table reveals two forms of attenuation. The conflict gap contracts across receiving scales, so fragile and robust trajectories induce increasingly similar update geometry. The negative rank association also weakens, so conflict gradually stops ordering which trajectories a student can continue. The fixed-8B rows expose both effects under one teacher distribution. Their agreement with the recovery curves links the optimization measurement to a concrete behavioral change.

### 7.4 Prefix Length Reveals How Recovery Fails

The intervention also clarifies fragility. The robust–fragile recovery gap is 65.65 percentage points after revealing 25% of a solution, 48.90 points after 50%, and 23.35 points after 75% (Figure 4). Fragile trajectories require substantially more correct reasoning before the student reaches a completable state. Conflict correlations follow the same pattern at every prefix fraction and approach zero at 4B.

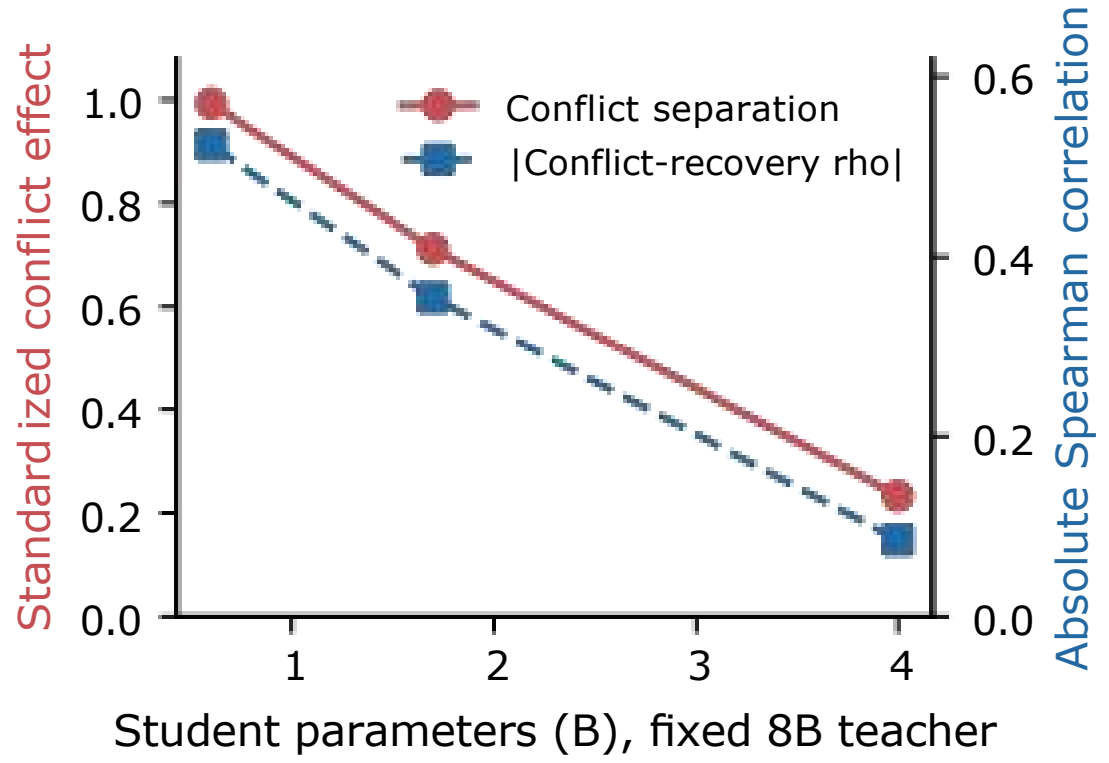

Figure 3: With the teacher fixed at 8B, both conflict separation and conflict–recovery coupling weaken as student capacity grows.

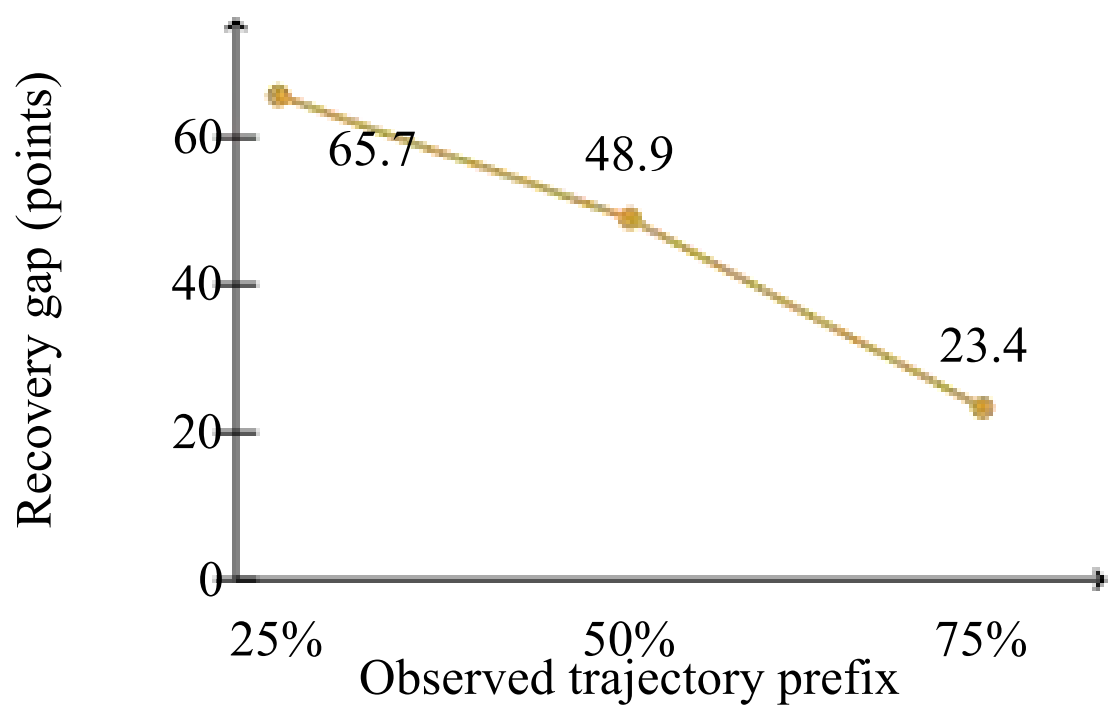

Figure 4: Difference in prefix recovery between robust and fragile trajectories as progressively longer prefixes are revealed.

The dose response localizes the missing support. A 25% prefix leaves much of the reasoning path to the student and produces the largest cohort separation. At 50%, the gap remains substantial, showing that many fragile trajectories contain difficult transitions beyond their opening steps. At 75%, the gap contracts sharply because the supplied context carries the student close to the terminal computation. The trajectories are thus fragile through their continuation structure, with recoverability increasing as verified reasoning bridges more of that structure.

## 8 Mechanistic Controls

Fixing the teacher at 8B isolates recipient capacity. Standardized conflict separation falls from 0.993 for the 0.6B student to 0.712 for 1.7B and 0.233 for 4B; the corresponding recovery correlations weaken from —0.5240 to —0.3547 and —0.0848. Changing the receiving model reproduces the attenuation in Figure 3.

With the student held fixed, changing the teacher from 4B to 8B preserves trajectory conflict ranks: Spearman $\rho$ = 0.9308 for 0.6B and 0.6516 for 1.7B. The receiving student determines much of which trajectories induce difficult updates.

Teacher scale still changes the magnitude of the distributional target. The stable rank ordering shows that this magnitude acts on a trajectory landscape largely organized by the student. Small recipients repeatedly encounter the same difficult states across teachers; larger recipients flatten that landscape as those states become reachable. This factorial pattern connects teacher information to student usability without reducing either side to a single model-level score.

The prefix intervention separates recoverability from static difficulty. At 0.6B, conflict–recovery correlations are —0.4625, —0.4713, and —0.3478 after 25%, 50%, and 75% prefixes. They weaken to —0.2614, —0.2956, and —0.2331 at 1.7B, then to —0.0352, —0.1060, and —0.0910 at 4B. The trend spans all three intervention points. In a five-fold cross-validated audit, posterior success and trajectory length explain $R^2$ = 0.6033 of recovery variation; objective conflict raises the explained variation to 0.6499.

The regression audit evaluates incremental signal at held-out trajectory identities. Posterior success captures how readily the untuned student already recognizes the answer region, and trajectory length captures a basic complexity proxy. Objective conflict adds information about the direction of supervision at the supplied state. The increase in held-out explained variation places conflict between a static difficulty feature and a complete predictor: it contributes a distinct optimization signal that becomes weak as capacity repairs the underlying continuation states.

Raw OlympiadBench accuracy ranges from 21.17% at 0.6B to 49.91% at 8B, leaving substantial error at every scale. The mathematical rKL improvement reaches zero at 4B. HumanEval gains follow the same attenuation from 3.66 to 2.44, 1.83, and 0.61 points. The training-trajectory recovery and conflict measurements locate this plateau before benchmark accuracy approaches a ceiling.

These controls align at different units of analysis. Benchmark error measures remaining task capability. Prefix recovery measures whether verified intermediate states enter the student's usable continuation set. Objective conflict measures how CE and reverse KL act at those states. Together, recovery heterogeneity and conflict coupling identify the student scales where distribution matching delivers its strongest cross-domain gains.

The $2 \times 2$ cohort–objective intervention tests the predicted direction. Reverse KL improves StepHard accuracy by 15.76 points over CE for fragile trajectories and by 8.79 points for stable trajectories. The resulting 6.97-point interaction, together with a 4.25-point mean OOD rescue for the fragile cohort, provides convergent mechanistic evidence.

# 9 Discussion

The three views of the experiment converge on a coherent and actionable explanation for diminishing gains: as student capacity grows, it recovers more verified trajectories from partial reasoning, the robust–fragile gap narrows, and CE–rKL conflict progressively loses its association with failure. The fixed-teacher comparison, combined with unsaturated OlympiadBench accuracy, pinpoints the transition as student-driven—usable trajectory support expands directly with the model's receptive capacity. This finding establishes recovery as a fundamental diagnostic of student capability, not merely a peripheral performance metric.

Prefix recovery elevates this mechanism into a powerful pre-training audit tool—one that transforms how practitioners evaluate and select synthetic reasoning data. By freezing verified traces, measuring recovery across multiple prefix fractions, and computing conflict on response tokens, practitioners gain a comprehensive diagnostic profile of their student model before committing to full-scale training. Broad recovery variation paired with negative conflict coupling provides strong justification for distribution matching. A residual fragile tail offers a targeted signal for decomposition or additional intermediate supervision. Uniformly high recovery, conversely, signals that investments are better directed toward new trajectories, retrieval, or alternative reasoning interfaces—enabling efficient resource allocation.

Beyond auditing, prefix recovery fundamentally enhances data curation. Low-recovery traces can be rewritten or decomposed to increase their pedagogical value; high-recovery traces can adopt simpler sequence training without compromising effectiveness. Critically, recovery injects student compatibility into existing filters based solely on final correctness and teacher confidence, creating a unified criterion that bridges teacher selection, trajectory design, and training objective. This integration offers a principled, model-aware approach to synthetic data evaluation that adapts to the learner's needs rather than applying one-size-fits-all quality thresholds. The framework thus provides both diagnostic clarity and practical leverage—equipping practitioners to make informed decisions about when and how to deploy distribution-matching supervision, ultimately unlocking greater returns from fixed training budgets.

# 10 Limitations

The current audit measures conditional continuation under fixed prefix fractions, decoding settings, verification protocols, and trajectory sources. The observed capacity boundaries may shift with different architectures, trace generators, or transfer objectives, and recovery summarizes behavioral success without resolving the internal representation of reasoning states. Several extensions merit future investigation. Denser prefix sampling could replace the current three-point summary with a full recovery curve, enabling more precise identification of where trajectories become self-sustaining. Token-level conflict analysis could be complemented with representation-level probes to examine how local signals propagate through complete solutions. Broader task families—including additional languages, tool-use settings, and interactive reasoning—would further validate the mechanism under diverse output constraints. These constitute natural next steps for building on the present foundation, and we treat them as open directions rather than limitations of the framework itself.

## 11 Conclusion

This work establishes prefix recovery as a principled and practical diagnostic for reasoning distillation. The core insight is that effective distillation depends not solely on the correctness or quality of teacher trajectories, but critically on whether the student can continue a verified solution from intermediate reasoning states. Prefix recovery measures this property directly, offering a student-conditioned alternative to conventional teacher-centered filtering criteria.

Across four adjacent model scales, the convergence of recovery patterns, objective conflict, and downstream transfer supports a coherent capacity-dependent account. Reverse KL emerges as the most broadly effective objective for the 0.6B and 1.7B students, precisely where correct trajectories exhibit the greatest heterogeneity in recoverability. As capacity increases, those trajectories become more uniformly recoverable, the conflict signal attenuates, and the marginal benefit of distributional supervision correspondingly diminishes—a consistent pattern that validates recovery as a predictor of transfer headroom.

Beyond explaining observed scaling behavior, prefix recovery provides a practical, pre-training diagnostic toolkit. Before committing to full-distribution training, practitioners can estimate transfer potential, identify trajectories where one-hot and distributional supervision diverge, and audit synthetic reasoning data without requiring process labels. This framework reframes distillation design by decoupling teacher strength from student compatibility—enabling informed decisions about when distribution matching is warranted, which trajectories merit priority, and where simpler supervision suffices. The result is a more efficient and model-aware approach to leveraging synthetic reasoning data, with immediate applicability to curriculum design and data curation in resource-constrained settings.

# A Experimental Details

**Models and datasets.** The main experiments use four adjacent Qwen3 teacher–student pairs. The two smaller students are fully tuned, whereas the 4B and 8B students use LoRA because full-parameter optimization exceeds the fixed hardware budget. Teacher parameters remain frozen in every arm. Mathematical transfer uses GSM8K trajectories and OlympiadBench evaluation; code transfer uses MBPP training examples and HumanEval evaluation. Training and test sources are disjoint.

| Domain | Train → Test | Train size | Test size |
|---|---|---|---|
| Mathematics | GSM8K → OlympiadBench | 7,473 | 581 |
| Code | MBPP → HumanEval | 120 | 164 |

Table 3: Primary training and evaluation data. All reported test results use the complete planned split.

**Optimization configuration.** All transfer arms use 800 optimizer steps, seed 42, temperature 1.0, learning rate $10^{-5}$, effective batch size 16, and maximum sequence length 1,024. Exact reverse KL is evaluated over the complete vocabulary on valid response tokens. Decoding settings and verifiers are fixed within every paired comparison.

| Student | Teacher | Update | Steps |
|---|---|---|---|
| 0.6B | 1.7B | Full | 800 |
| 1.7B | 4B | Full | 800 |
| 4B | 8B | LoRA | 800 |
| 8B | 14B | LoRA | 800 |

Table 4: Frozen adjacent-scale training configuration.

**Predictions and identifying controls.** The main study tests three linked predictions under paired experi - mental units. Table 5 records the quantity and control associated with each prediction; it is included here because it documents the identification design rather than reporting a primary outcome.

| Prediction | Experimental design |
|---|---|
| Fragile trajectories are repaired first | Trajectory-level prefix recovery; paired 0.6B, 1.7B, and 4B responses; identical trajectories and prefix fractions. |
| Conflict matters less with capacity | Token-level CE–rKL conflict; robust–fragile gap and ρ(C, R); fixed 8B teacher and full-vocabulary gradients. |
| Transfer headroom contracts | Item-level correctness change; adjacent rKL versus the raw student; fixed recipe, verifier, decoding, and complete split. |

Table 5: Linked predictions and controls used to distinguish student capacity from teacher identity, trajectory composition, and evaluation noise.

# B Prefix–Recovery Analysis

**Cohort construction and generation.** The recovery cohort contains 500 fragile and 500 robust verified trajectories. Cohorts are selected once from low and high posterior success and matched by realized trajectory length before prefix generation. These labels describe behavior for the smallest student; they are not claims that a question is intrinsically or permanently fragile. Reusing the same question identifiers and trajectories across scales allows the analysis to measure which states larger students repair.

For each trajectory, response tokens are truncated at 25%, 50%, and 75%. Four continuations are sampled at each fraction, yielding 12 observations per trajectory and student. The 0.6B, 1.7B, and 4B studies therefore contain 36,000 real continuations in total. A continuation is marked correct only when the repository's robust verifier accepts the reconstructed answer. Formal analysis rejects incomplete cohorts, duplicate question identifiers, mismatched trajectory hashes, or record counts that disagree with the frozen manifest.

| Student | Mean recovery | Fragile | Robust |
|---|---|---|---|
| 0.6B | 70.97% | 47.98% | 93.95% |
| 1.7B | 85.45% | 72.53% | 98.37% |
| 4B | 91.93% | 84.75% | 99.10% |

Table 6: Prefix recovery on the fixed 1,000-trajectory cohort.

**Recovery results.** The paired recovery gain is 14.48 percentage points from 0.6B to 1.7B and 6.48 points from 1.7B to 4B. The robust–fragile difference contracts by 31.62 points between 0.6B and 4B. At the individual prefix fractions, the 0.6B conflict–recovery correlations are −0.4625, −0.4713, and −0.3478; the 1.7B correlations are −0.2614, −0.2956, and −0.2331; and the 4B correlations are −0.0352, −0.1060, and −0.0910. Thus the capacity trend does not depend on selecting one favorable prefix length.

# C Additional Mechanistic Controls

**Exact objective–conflict computation.** For student distribution $p = \mathrm{softmax}(z)$, teacher distribution $q$, and one-hot target $e_y$, the CE logit gradient is $g_{CE} = p - e_y$. If $D = D_{KL}(p \,\|\, q)$, the reverse-KL gradient at vocabulary coordinate $i$ is

$$g_{rKL,i} = p_i(\log p_i - \log q_i - D) \,. \tag{5}$$

Objective conflict is $(1 - \cos(g_{CE}, g_{rKL}))/2$ and is averaged over valid response tokens. Both gradients are computed over the complete vocabulary; no top-k approximation is used.

To separate student capacity from teacher identity, conflict is computed for 0.6B/4B, 0.6B/8B, 1.7B/4B, 1.7B/8B, and 4B/8B student–teacher pairs. The fixed-8B comparison changes only the student. Standardized robust–fragile separation decreases from 0.993 to 0.712 and 0.233 across the 0.6B, 1.7B, and 4B students. When the student is fixed and the teacher changes from 4B to 8B, trajectory conflict ranks remain correlated at 0.9308 for the 0.6B student and 0.6516 for the 1.7B student.

| Student/teacher | Conflict gap | Std. effect | ρ(C, R) |
|---|---|---|---|
| 0.6B/4B | 0.0370 | 0.968 | -0.5016 |
| 0.6B/8B | 0.0446 | 0.993 | -0.5240 |
| 1.7B/4B | 0.0210 | 0.615 | -0.3092 |
| 1.7B/8B | 0.0234 | 0.712 | -0.3547 |
| 4B/8B | 0.0064 | 0.233 | -0.0848 |

Table 7: Complete full-vocabulary objective-conflict analysis.

**Prefix–length and covariate controls.** Trace length and initial posterior success can correlate with difficulty, but neither changes the amount of reasoning supplied to the student. Prefix recovery exhibits a controlled dose response: the robust–fragile difference falls from 65.65 percentage points at 25% context to 48.90 points at 50% and 23.35 points at 75%. In a five-fold cross-validated regression audit, adding objective conflict to posterior success and trajectory length increases $R^2$ from 0.6033 to 0.6499. This comparison supports incremental diagnostic information without claiming that conflict is a sufficient training rule.

# D Mathematical Derivations

This section derives the quantities used in the empirical analysis. The results establish what the estimators measure and how pairing preserves the experimental correspondence; they do not assume that model capacity itself is randomized.

**Exact reverse–KL geometry.** Let $a_i = \log p_i - \log q_i$ and assume $p_i, q_i > 0$, as holds for finite softmax logits. Then $D_{KL}(p\|q) = \sum_j p_j a_j$.

**Proposition 1** (Exact rKL gradient)**.** *For* $p = \text{softmax}(z)$*,*

$$\frac{\partial D_{\mathrm{KL}}(p\|q)}{\partial z_i} = p_i\left(a_i - D_{\mathrm{KL}}(p\|q)\right). \tag{6}$$

*Moreover,* $\sum_i \partial D_{KL}(p\|q)/\partial z_i = 0$*.*

*Proof.* The softmax Jacobian is $\partial p_j/\partial z_i = p_j(\mathbf{1}\{i = j\} - p_i)$. Differentiating $\sum_j p_j (\log p_j - \log q_j)$ therefore gives

$$\frac{\partial D}{\partial z_i} = \sum_j p_j(\mathbf{1}\{i=j\} - p_i)(a_j + 1) \tag{7}$$

$$= p_i(a_i + 1) - p_i \sum_j p_j(a_j + 1) \tag{8}$$

$$= p_i(a_i - D), \tag{9}$$

because $\sum_j p_j = 1$ and $\sum_j p_j a_j = D$. Summing the final expression over i yields $D - D = 0$. Thus the gradient lies in the logit simplex's tangent space and is invariant to adding a common constant to all logits.

**Proposition 2** (Range and interpretation of objective conflict)**.** *For nonzero gradients* $u = g_{CE}$ *and* $v = g_{rKL}$*, define*

$$C(u, v) = \frac{1 - \langle u, v\rangle/(\|u\|_2\|v\|_2)}{2}. \tag{10}$$

*Then* $0 \leq C(u, v) \leq 1$*. The endpoints correspond respectively to positive collinearity and negative collinearity, while orthogonal gradients give* $C = 1/2$*.*

*Proof.* Cauchy–Schwarz gives $-1 \leq \langle u, v\rangle/(\|u\|_2\|v\|_2) \leq 1$. Applying the affine map $x \mapsto (1 - x)/2$ proves the bound and the three stated cases. Tokens with a zero gradient have no defined direction and are excluded before trajectory averaging.

These propositions also clarify why the paper computes both gradients over the complete vocabulary. Renormalizing a top-k subset changes p, changes the expectation D in Equation 6, and can therefore change both the magnitude and direction of the measured conflict.

**Recovery as a paired Monte Carlo estimand.** Let F be the fixed set of prefix fractions and let $V_{sfk}(y) \in \{0, 1\}$ denote verifier success for continuation k from trajectory y, student s, and fraction f. Write $\pi_{sf}(y) = \Pr[V_{sfk}(y) = 1 \mid y, f, s]$ and

$$\widehat{R}_s(y) = \frac{1}{|\mathcal{F}|K} \sum_{f \in \mathcal{F}} \sum_{k=1}^{K} V_{sfk}(y). \tag{11}$$

**Proposition 3** (Recovery estimator). $\widehat{R}_s(y)$ *is unbiased for* $R_s(y) = |F|^{-1}\Sigma_f \pi_{sf}(y)$. *If continuations conditionally independent given* $(y, f, s)$*, then*

*are*

$$\begin{aligned}\operatorname{Var}[\widehat{R}_s(y) \mid y] &= \frac{1}{|\mathcal{F}|^2 K} \sum_{f} \pi_{sf}(y)(1 - \pi_{sf}(y)) \\ &\le \frac{1}{4|\mathcal{F}|K}.\end{aligned} \tag{12}$$

*For the study's* $|F| = 3$ *and* $K = 4$*, the upper bound is* 1/48.

*Proof.* Linearity of expectation gives $\mathbb{E}[\widehat{R}_s(y) \mid y] = (|\mathcal{F}|K)^{-1} \sum_f K\pi_{sf}(y) = R_s(y)$. Conditional independence makes the variance of the sum equal the sum of Bernoulli variances, yielding Equation 12. Finally, $x(1-x) \le 1/4$ for $x \in [0, 1]$.

The unbiasedness statement does not require monotonic recovery as a longer prefix is revealed. Indeed, monotonicity is not built into the metric and is treated as an empirical dose-response question. The fixed fractions instead define the same finite intervention distribution for every student.

**Proposition 4** (Gap-closure identity). *Let* $G \in \{r, f\}$ *denote the frozen robust and fragile cohorts, let* $\bar{R}_{s,G} = |G|^{-1}\Sigma_{y \in G} R_s(y)$*, and define the cohort gap* $B_s = \bar{R}_{s,r} - \bar{R}_{s,f}$*. For two students* s *and* s′*, define within-cohort gains* $\Gamma_G = \bar{R}_{s',G} - \bar{R}_{s,G}$*. Then*

$$B_s - B_{s'} = \Gamma_f - \Gamma_r. \tag{13}$$

*Consequently, a positive gap closure is exactly equivalent to larger average repair of the fragile cohort than of the robust cohort.*

*Proof.* Expanding and regrouping gives

$$B_s - B_{s'} = (\bar{R}_{s,r} - \bar{R}_{s,f}) - (\bar{R}_{s',r} - \bar{R}_{s',f}) \tag{14}$$

$$= (\bar{R}_{s',f} - \bar{R}_{s,f}) - (\bar{R}_{s',r} - \bar{R}_{s,r}) = \Gamma_f - \Gamma_r. \tag{15}$$

The identity is algebraic and does not depend on equal cohort sizes, although the experiment uses 500 trajectories in each cohort.

**Why item and trajectory pairing matters.** Let $A_i$ and $B_i$ be two binary outcomes observed on the same evaluation item, and let $D_i = A_i - B_i$.

**Proposition 5** (Variance of a paired difference). *For independent items,*

$$\operatorname{Var}\left(\frac{1}{N}\sum_i D_i\right) = \frac{\operatorname{Var}(A) + \operatorname{Var}(B) - 2\operatorname{Cov}(A, B)}{N}. \tag{16}$$

*An analysis that discards pairing omits the covariance term. Hence pairing is at least as efficient whenever item difficulties induce* $\mathrm{Cov}(A, B) \geqslant 0$*, and its gain is exactly* $2\mathrm{Cov}(A, B)/N$.

*Proof.* Equation 16 follows from $\mathrm{Var}(A-B) = \mathrm{Var}(A)+\mathrm{Var}(B)-2\mathrm{Cov}(A, B)$ and independence across items. Removing the item correspondence produces the sum of the two marginal variance terms, so subtracting the paired variance yields the stated gain.

For binary outcomes, the observed transfer effect can also be written as $\bar{\Delta} = (n_{10} - n_{01})/N$, where $n_{10}$ counts items solved only by the transfer model and $n_{01}$ counts items solved only by the baseline. Items on which both models agree contribute zero. Retaining this correspondence preserves the discordant pairs. Recovery comparisons similarly keep the measurements belonging to one trajectory together rather than treating its 12 continuations as 12 independent experimental units.

# E From Measurement to Training Decisions

Prefix recovery redefines the evaluation of synthetic reasoning data by shifting the unit of assessment from the trace itself to the student model's interaction with it. Standard filtering judges whether a trace is correct, optionally augmented by teacher confidence or length penalties—criteria that are teacher- or example-centered, applying the same decision to all students regardless of their individual capabilities. Recovery, in contrast, is student-centered. It measures how much of the reasoning must be provided before the student can reliably complete the solution, while preserving the trajectory identity throughout the evaluation.

This framework offers fine-grained diagnostic power. A low-recovery trace can arise from two operationally distinct causes: the student may lack foundational concepts needed to proceed from any intermediate point, or it may recognize the solution only after a critical transformation has been exposed. Analyzing recovery across multiple prefix fractions disentangles these cases. Persistently low recovery across all prefixes indicates the trace lies outside the student's current reachable set, while a steep late recovery spike pinpoints a specific missing transition—directing supervision to expose or decompose that bottleneck.

The framework further incorporates objective conflict as a second diagnostic dimension. Low recovery accompanied by minimal CE–rKL conflict points to a representation or search limitation shared across objectives. Conversely, low recovery with high conflict identifies decision points where the choice of supervision materially alters the local update direction—precisely the regime where distribution matching can provide concrete benefit by retaining teacher-supported alternatives that one-hot imitation would suppress.

These measurements enable a principled and conservative decision procedure. Practitioners first freeze a verified cohort representative of the intended training source, independent of the target benchmark. Recovery is measured at multiple prefix fractions for the candidate student, and conflict is computed against candidate teachers on the same response tokens. Full-distribution training is justified when recovery remains heterogeneous across traces and low-recovery examples exhibit systematic conflict. Conversely, uniformly high recovery with weak coupling to conflict suggests that cheaper sequence supervision or direct task training may suffice, providing a clear cost-benefit rationale for practitioners.

The procedure deliberately avoids learned thresholds, as fitting one to OlympiadBench or HumanEval would entangle diagnosis with the outcomes it is meant to predict. Recovery, the robust–fragile contrast, and conflict association are therefore maintained as independent diagnostics, with their capacity trends compared against independently measured downstream transfer. A future selector should fix its rule on one source and validate it on an untouched model family and benchmark—ensuring generalizability rather than overfitting to specific evaluation conditions.

The computational efficiency of this approach is noteworthy. Prefix generation requires only inference from the untuned student, and teacher-gradient evaluation demands no optimizer trajectory. Both analyses can be executed on a bounded cohort before committing the substantially larger budget required for complete distillation and downstream evaluation. The same measurements also specify a testable and actionable curriculum: prioritize traces that become recoverable after a moderate prefix, gradually shorten the supplied context over training to reinforce learning, and decompose or defer traces that remain fundamentally unrecoverable—transforming diagnostic insights directly into training strategy.

# F Statistical Interpretation and Falsifiability

The primary downstream quantities are paired accuracy differences, not independent model accuracies. For benchmark item i, the contribution is $c_i^{rKL} - c_i^{base}$ under identical decoding and verification. Retaining item correspondence records whether the two models succeed on the same question. Recovery comparisons analogously pair the same trajectory across capacities. This avoids treating the twelve continuations of one trajectory as twelve independent benchmark items.

The evidence has different resolution at different levels. HumanEval contains 164 items, whereas the recovery analysis pairs 1,000 trajectories. The cross-domain statement is therefore descriptive sign replication at two sub-2B scales, while the objective intervention is supporting mechanistic evidence rather than a separate universal benchmark claim.

Several observations would contradict the proposed account. Stable conflict separation under the fixed-8B teacher would leave teacher identity as a sufficient explanation. Fragile trajectories remaining fragile as students grow would contradict expansion of the reachable continuation set. Growing rKL gains after recovery variation and conflict coupling disappear would show that the diagnostic does not track transfer headroom. None occurs in the evaluated scales: all three signals attenuate, although at different rates.

The claim would also be weakened by a domain reversal below 2B. Mathematics and code are therefore reported separately rather than averaged, reweighted by sample count, or selected according to effect size. The relevant observation is that both changes are positive at 0.6B and 1.7B and no longer consistently positive at larger scales.

Evaluation integrity provides a final guard against accidental confirmation. Every reported benchmark uses its complete planned split, item identifiers are retained for pairing, and records are rejected when manifest counts and verifier outputs disagree. The trajectory cohort is reused across scales rather than regenerated after observing results.

Taken together, the study is a triangulated capacity intervention. Downstream transfer supplies external validity, prefix recovery supplies a behavioral mechanism, and objective conflict supplies an optimization correlate. No single layer proves causality on its own. Their coupled attenuation under paired trajectories and a fixed-teacher control is the central evidence, and the explicit failure conditions define what a replication must test.